\def\year{2018}\relax
\documentclass[letterpaper]{article} 
\usepackage{aaai18}  
\usepackage{times}  
\usepackage{helvet}  
\usepackage{courier}  
\usepackage{url}  
\usepackage{amsmath}
\usepackage[noend]{algorithmic}
\usepackage[ruled]{algorithm2e}
\usepackage{amssymb}
\usepackage{CJKutf8}
\usepackage{CJK}
\usepackage{epstopdf}
\usepackage{times}
\usepackage{makecell}
\usepackage{url}
\usepackage{latexsym}
\usepackage{graphicx}
\usepackage{amsmath}
\usepackage{amssymb}
\usepackage{amsmath}
\usepackage{CJK}
\usepackage{graphicx}  
\usepackage[usenames, dvipsnames]{color}
\usepackage{tabulary}
\newlength\myindent
\newcolumntype{C}[1]{>{\centering\let\newline\\\arraybackslash\hspace{0pt}}m{#1}}
\newcolumntype{L}[1]{>{\raggedright\arraybackslash}p{#1}}
\begin{document}
	\begin{CJK*}{UTF8}{gbsn} 
		%
		\title{Dictionary-Guided Editing Networks for Paraphrase Generation}
\author{
	Shaohan Huang$^\dag$~, Yu Wu$^\ddag$, Furu Wei$^\dag$,	Ming Zhou$^\dag$~~~~\\
	$^\dag$Microsoft Research, Beijing, China\\
    $^\ddag$State Key Lab of Software Development Environment, Beihang University, Beijing, China\\
	\{shaohanh, fuwei, mingzhou\}@microsoft.com wuyu@buaa.edu.cn 
    }
		\maketitle
		\begin{abstract}
An intuitive way for a human to write paraphrase sentences is to replace words or phrases in the original sentence with their corresponding synonyms and make necessary changes to ensure the new sentences are fluent and grammatically correct. We propose a novel approach to modeling the process with dictionary-guided editing networks which effectively conduct rewriting on the source sentence to generate paraphrase sentences. It jointly learns the selection of the appropriate word level and phrase level paraphrase pairs in the context of the original sentence from an off-the-shelf dictionary as well as the generation of fluent natural language sentences. 
Specifically, the system retrieves a set of word level and phrase level paraphrased pairs derived from the Paraphrase Database (PPDB) for the original sentence, which is used to guide the decision of which the words might be deleted or inserted with the soft attention mechanism under the sequence-to-sequence framework. 
We conduct experiments on two benchmark datasets for paraphrase generation, namely the MSCOCO and Quora dataset. The evaluation results demonstrate that our dictionary-guided editing networks outperforms the baseline methods.
		\end{abstract}
        
\section{Introduction}
Paraphrase generation aims to generate restatements of the meaning of a text or passage using other words. 
It is a fundamental task in natural language processing with many applications in information retrieval, question answering, dialogue, and conversation systems. Existing work on paraphrase generation focuses on generating paraphrase sentences from scratch. For example, \citeauthor{lin2014microsoft} propose generating paraphrases with statistical machine translation models. Recently, neural networks based generative models under the sequence-to-sequence framework have also been used for paraphrase generation~\cite{lin2014microsoft}.
 
 \begin{figure}[h]		
			\begin{center} 
				\includegraphics[width=8.5cm,height=3.5cm]{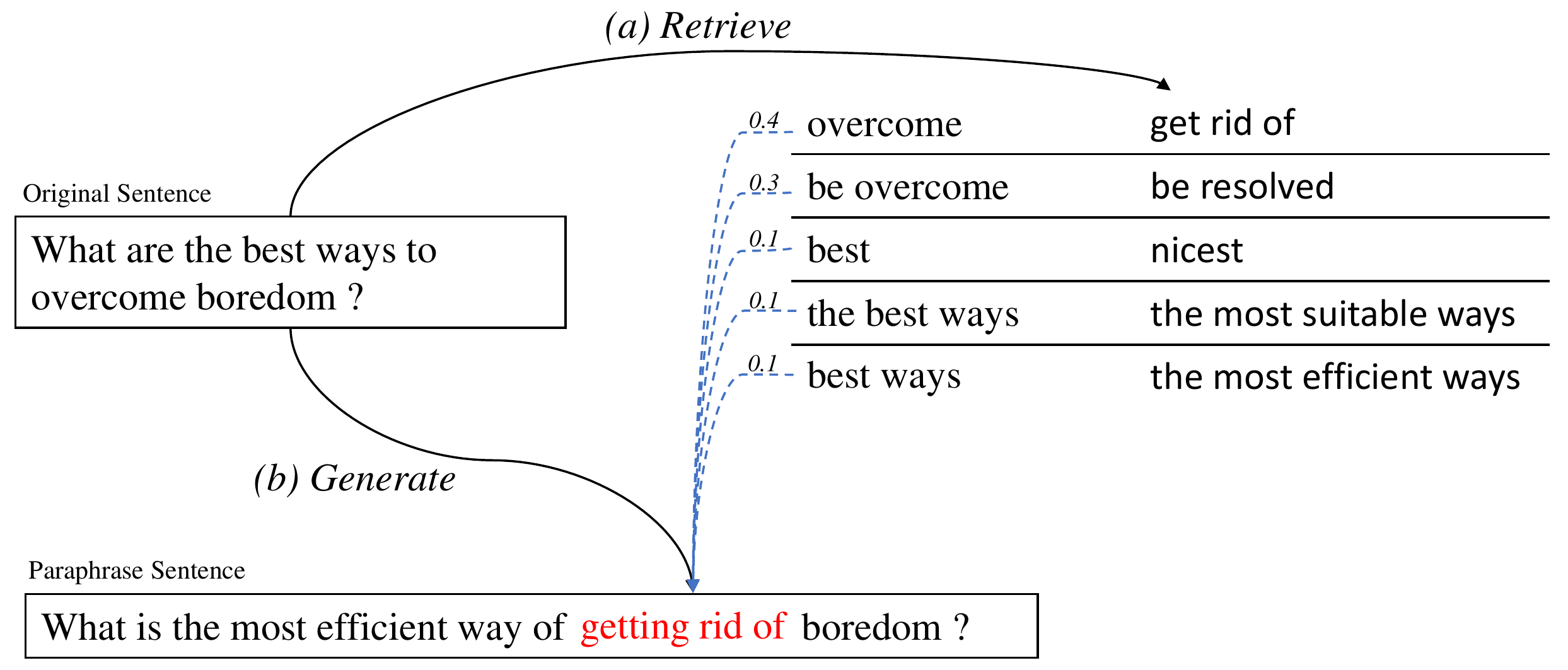}
			\end{center}
			\vspace{-2mm}
			\caption{The dictionary-guided editing networks model first retrieves a group of paraphrased pairs and then generates a paraphrase using the original sentence as a prototype.
			}\label{fig:over}
		\end{figure}

However, an intuitive way for a human to write paraphrase sentences is to replace words or phrases in the original sentence with their corresponding synonyms and make necessary changes to ensure the new sentences are fluent and grammatically correct. Figure~\ref{fig:over} shows an example. Given the input sentence ``What are the best ways to overcome boredom?'', we can first replace ``overcome'' with the word level paraphrase phrases ``get rid of'', and then make small changes over the new sentence to ensure it is grammatically correct and fluent. Certainly, it should be emphasized that the selection of context-relevant paraphrase pairs from an off-the-shelf dictionary with respect to the original sentence is also important for a good revision. This process demonstrates that humans usually write paraphrase sentences by editing the input sentence, which motivates us to develop models for paraphrase generation through editing.

We are inspired by \citeauthor{gupta2017deep}'s pioneer work on a new paradigm to generate sentences. Specifically, they propose a new generative model of sentences that first samples a prototype sentence from the training corpus and then edits it into a new sentence. Unlike randomly sampling the edit vector to generate a new sentence, we can leverage the off-the-shelf word level and phrase level paraphrase pairs (e.g. synonyms) to construct the editing vector where the deletion of words from the original sentence and the insertion words into the target sentence can be explicitly modeled.

In this paper, we propose a dictionary-guided editing networks for paraphrase generation which effectively conducts rewriting on the source sentence to generate paraphrase sentences. It jointly learns the selection of the appropriate word level and phrase level paraphrase pairs in the context of the original sentence from an off-the-shelf dictionary as well as the generation of fluent natural language sentences. The system retrieves a set of word level and phrase level paraphrased pairs derived from the Paraphrase Database (PPDB) for the original sentence, which are used to guide the decision on which the words might be deleted or inserted with the soft attention mechanism under the sequence-to-sequence framework. 

We conduct experiments on the benchmark MSCOCO and Quora datasets for paraphrase generation. The evaluation results demonstrate that the dictionary-guided editing networks outperforms existing sequence-to-sequence generation baselines and achieves state-of-the-art results.

        
The rest of this paper is organized as follows: Section 2 gives a brief overview of the recent history of paraphrase generation and presents a description of text editing methods. In Section 3 we show the detailed design of our dictionary-guided editing network model. In Section 4 we conduct paraphrase generation experiments on two datasets and demonstrate the  evaluation results. Section 5 concludes this paper and outlines future work.

\section{Related Work}
Paraphrase generation aims to generate a semantically equivalent sentence with different expressions. Prior approaches  can be categorized into knowledge-based approaches and statistical machine translation (SMT) based approaches.
Knowledge-based approaches primarily rely on hand-crafted rules and dictionaries that enjoy high precision but that are hard to scale up. The pioneer of this approach is Kozlowski et al. \shortcite{kozlowski2003generation} who first pair simple semantic structures with their syntactic realization and then generate paraphrases using such predicate/argument structures. A famous paraphrase generation system is designed by Hassan et al. \shortcite{hassan2007unt}, where paraphrases are generated by word substitutions and the substitution table is obtained by leveraging several external resources, such as WordNet and Microsoft Encarta encyclopedia. Subsequently, Madnani and Dorr \shortcite{madnani2010generating} propose a knowledge-driven method by using hand crafted rules or automatically learned complex paraphrase patterns \cite{zhao2009application}.  SMT based paraphrase generation is proposed by  \cite{quirk2004monolingual}, where an SMT model is trained on large volumes of sentence pairs extracted from clustered news articles. Zhao et al. \shortcite{zhao2008combining} combine multiple resources to learn phrase-based paraphrase tables and corresponding feature functions to devise a log-linear SMT model. To leverage the power of multiple machine translate engine, a multi-pivot approach is proposed in \cite{zhao2010leveraging}  to obtain plenty of paraphrase candidates. Then these candidates are used by selection-based and decoding-based methods to produce high-quality paraphrases.

Recently, deep learning-based approaches have been introduced for paraphrase generation and achieved great success. Prakash et al. \shortcite{prakash2016neural} employ the residual recurrent neural networks for paraphrase generation, that is one of the first major words that uses a deep learning model for this task. Gupta et al. \shortcite{gupta2017deep} propose a combination of variational autoencoder(VAE) and sequence-to-sequence model to generate paraphrase. We also investigate deep learning for paraphrase generation, and we are the first one to utilize an editing mechanism for this task. 

Finally, our work is in the spirit of prototype editing methods for natural language generation \cite{DBLP:journals/corr/abs-1709-08878}, which proposes a generative model that first samples a prototype sentence from training data and then edits it into a new sentence. We utilize the original sentence as a prototype and learn the edit vector from paraphrase dataset (PPDB) \cite{ganitkevitch2013ppdb}. Li et al. \shortcite{li2018delete} introduce a simple approach for style transfer. It can be considered for applying content words by deleting phrases associated with original attribute values as a prototype, and combining a new phrase with the target attribute to generate a final output. Cao et al. \shortcite{ziqiang2018summary} employ existing summaries as soft templates, and rerank these soft templates by considering the current document. Finally, a summary is generated with a seq2seq framework augmented with the templates. Our work can be seen as an extension of editing methods for paraphrase generation. The stark difference is that our model is capable of leveraging an external dictionary in editing, which ensures that the expression changes do not affect its original semantic.

\section{Methodology}

\begin{figure*}[t]		
			\begin{center}
				\includegraphics[width=14cm,height=7.3cm]{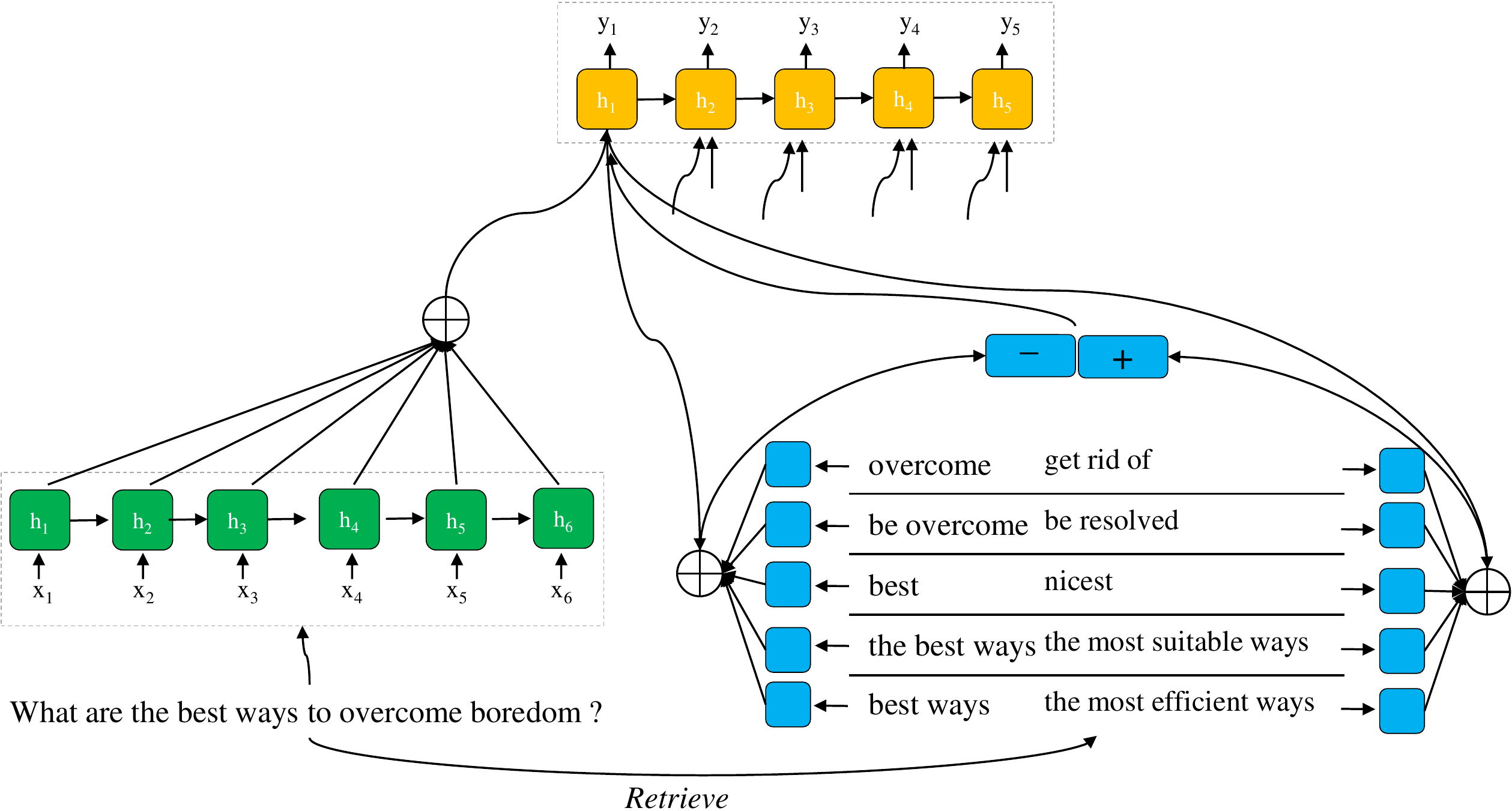}
			\end{center}
			\caption{Architecture of dictionary-guided editing networks. At each step of the decoder, we implement the soft attention mechanism to guide the decision for word deletion or insertion.}\label{fig:arch}
		\end{figure*}
		
\subsection{Problem Definition}


We assume there is access to a corpus of lexical or phrasal paraphrased dictionary $\mathcal{D} = \{(o_i,p_i)\}^N_{i=1} $, where $o_i$ is an original word or phrase and $p_i$ is the word-level or phrase-level paraphrase of $o_i$. Besides, we have a parallel data set $\mathcal{P} = \{(x_i,y_i)\}^L_{i=1}$, where $(x_i,y_i)$ is a paraphrase pair.  Our goal is to learn a paraphrase generator with the use of $\mathcal{D}$ and $\mathcal{P}$, so as to precisely paraphrase a new sentence $x$ with $y$.

The overview of our model is shown in Figure~\ref{fig:arch}. We first retrieve a set of word level or phrase level paraphrased pairs $\mathcal{E} = \{(o_i,p_i)\}^M_{i=1} $ where $ (o_i,p_i) \in \mathcal{D}$ for original sentence $x$. Secondly, we implement a neural encoder to convert each word or phrase into a vector in $\mathcal{E}$, which is used in the soft attention mechanism. Finally, we learn the dictionary-guided editing networks model to generate the paraphrase sentence $y$.

\subsection{Retrieve}
Our model relies on the observation that humans usually write paraphrase sentences by replacing words or phrases in the original sentence with their corresponding synonyms. Therefore, the first step of our method is to retrieve a set of lexical or phrasal paraphrased pairs dictionary for the original sentence. For example, for original sentence $x$ \textit{``What are the best ways to overcome boredom''}, we would find some paraphrased pairs such as (\textit{``overcome''}, \textit{``get rid of''}), (\textit{``the best ways''}, \textit{``the most suitable ways''}), and (\textit{``the best ways''}, \textit{``the most efficient ways''}).

Our system retrieves word level and phrase level paraphrased pairs derived from the Paraphrase Database (PPDB) \cite{pavlick2015ppdb}. PPDB is an automatically extracted database containing millions of paraphrases in different languages. It contains three types of paraphrases: lexical (single word to single word), phrasal (multiword to single/multiword), and syntactic (paraphrase rules containing non-terminal symbols). We use PPDB with the lexical and phrasal types as raw paraphrased dictionary $\mathcal{D}$.

We leverage Lucene\footnote{https://lucene.apache.org/} to index the paraphrased dictionary $\mathcal{D}$ and use the default ranking function in Lucene during the search phase. Specifically, we  index all $o_i$ in $\mathcal{D}$, and retrieve top $10 \times M$ paraphrased pairs for the original sentence $x$. Then, we  rank $10 \times M$ candidates by combining the TF-IDF weighted word overlap and the PPDB score. The ranking score is formulated as:
\begin{eqnarray} 
		score_r = \sum_{w \in o_i \cap x} tf_w \cdot idf_w + score'(o_i, p_i)
\end{eqnarray}
where $score'(o_i, p_i)$ is the PPDB score for $(o_i, p_i)$ pairs, which is computed by a regression model \cite{pavlick2015ppdb} in PPDB. We obtain a set of word level or phrase level paraphrased pairs $\mathcal{E}$ as the local dictionary for the original sentence $x$.

\subsection{Dictionary Encoder}
After finding a group of paraphrased pairs $\mathcal{E} = \{(o_i,p_i)\}^M_{i=1} $ for original sentence $x$, we use a neutral dictionary encoder to convert $\mathcal{E}$ into a representation vector. In the case of single word paraphrased pairs, a good representation vector would be the word vector of $o_i$ or $p_i$. For multiple words, $o_i$ or $p_i$ are represented as the sum of the individual word vectors \cite{gupta2017deep}.
\begin{eqnarray} \small
		o^i_r =  \sum_{w \in o_i}  \Phi(w) \\
        p^i_r =  \sum_{w \in p_i}  \Phi(w) 
\end{eqnarray}
where $\Phi(w)$ is the word vector for word $w$ and $o^i_r$ is the representation vector of $o_i$ and $p^i_r$ is the representation vector of $p_i$.

For each paraphrased pair in $\mathcal{E}$, we employ the same encoding method and obtain $2 \times M$ vectors $\mathcal{E'}=\{(o^i_r,p^i_r)\}^M_{i=1}$. In the next section, we will introduce leveraging our paraphrased dictionary to generate a paraphrase.

\subsection{Dictionary-Guided Editing}
We propose a dictionary-guided editing networks model where paraphrased dictionary is used to guide the decision for words that might be deleted or inserted with the soft attention mechanism under the sequence-to-sequence framework. We learn our model that takes as input original sentence $x$ and representation vectors $\mathcal{E'} = \{(o^i_r,p^i_r)\}^M_{i=1}$. 

For original sentence $x$, we first regard the output of the BiRNN as the representation of the original sentence $x$ and use the standard attention model \cite{luong2015effective} to capture original-side information. 

For representation vectors $\mathcal{E'}$, we adopt the soft attention mechanism, which is introduced to better utilize paraphrased dictionary information. The soft attention mechanism would be used to guide the decision for word deletion or insertion in each step of the decoder.

For the t-th time step of the decoder, $\mathbf{h_t}$ denotes its hidden state. The goal is to derive a context vector $\mathbf{c_t}$ that captures paraphrased dictionary side information to guide the decoder. We employ a concatenation layer to combine $\mathbf{h_t}$, $\mathbf{c_t}$ and $\mathbf{c'_t}$ as follows:
\begin{equation}
\mathbf{\tilde{h}_t} = tanh(\mathbf{W_c} \cdot (\mathbf{h_t} \oplus \mathbf{c_t} \oplus \mathbf{c'_t}))
\end{equation}
where $\oplus$ denotes concatenation and $\mathbf{W_c}$ is a parameter. The vector $\mathbf{c'_t}$ is the standard attention for the source side. $\mathbf{c'_t}$ is computed as the weighted average of the original hidden states.

We then compute context vector $\mathbf{c_t}$. In paraphrased pairs $\mathcal{E} = \{(o_i,p_i)\}^M_{i=1}$, $o_i$ might be the word that will be deleted and $p_i$ might be inserted. In order to better guide our decoder on which word might be deleted or be inserted, we employ two soft attentions to compute the $o_i$-side and $p_i$-side context vectors respectively. Context vector $\mathbf{c_t}$ is computed as the weighted average of $o^i_r$ and $p^i_r$ as follows:
\begin{equation}
\mathbf{c_t} = \sum_{i=1}^M \mathbf{a_{t, i}} \cdot o^i_r \oplus \sum_{i=1}^M \mathbf{a'_{t,i}} \cdot p^i_r
\end{equation}

The $\mathbf{a_t}$ and $\mathbf{a'_t}$ are alignment vectors, whose size equals $M$. $\mathbf{a_{t, i}}$ is formulated as:
\begin{eqnarray} \label{eq:del}
		 && \mathbf{a_{t, i}} = \frac{exp(score(\mathbf{h_t}, o^i_r))}{\sum_{j=1}^M exp(score(\mathbf{h_t}, o^i_r))} \\
		&& score(\mathbf{h_t}, o^i_r) = v^\top tanh(W_{\alpha}[\mathbf{h_t}  \oplus o^i_r])
\end{eqnarray}
where  $W_{\alpha}$ and $v$ are parameters. The $p^i_r$-side alignment vector $\mathbf{a'_{t, i}}$ is formulated as:
\begin{eqnarray} \label{eq:ins}
		 && \mathbf{a_{t, i}} = \frac{exp(score(\mathbf{h_t}, p^i_r))}{\sum_{j=1}^M exp(score(\mathbf{h_t}, p^i_r))} \\
		&& score(\mathbf{h_t}, p^i_r) = v'^\top tanh(W'_{\alpha}[\mathbf{h_t}  \oplus p^i_r])
\end{eqnarray}
where $W'_{\alpha}$ and $v'$ are the attention parameters. 

A softmax layer is introduced to compute probability distribution of the t-th time word:
		\begin{equation}
		\mathbf{y_t} = softmax(\mathbf{W_y} [\mathbf{y_{t-1}} \oplus \mathbf{\tilde{h}_t} \oplus \mathbf{c_t} \oplus \mathbf{c'_t}] +\mathbf{b_y})
		\end{equation}
		where $\mathbf{W_y} $ and $\mathbf{b_{y}}$ are two parameters. 
		
For the generative model, the learning goal is to maximize the probability of the actual paraphrase $\mathbf{y^*}$. We learn our model by minimizing the negative log-likelihood (NLL):
		\begin{equation} 
		\mathcal{J} =  -\text{log}(p(\mathbf{y^*}|\mathbf{x}, \mathcal{E'}))
		\end{equation}

The mini-batched Adam \cite{kingma2014adam} algorithm is used to optimize the objective function. In order to avoid overfitting, we adopt dropout layers between different LSTM layers same as \cite{zaremba2014recurrent}.

\section{Experiments}
\subsection{Datasets}
We present the performance of our model on two benchmark datasets, namely the MSCOCO and Quora datasets.

\textbf{MSCOCO} \cite{lin2014microsoft} is a large-scale captioning dataset which contains human annotated captions of over 120K images \footnote{http://cocodataset.org/}. This dataset was used previously to evaluate paraphrase generation methods \cite{prakash2016neural,gupta2017deep}. In the MSCOCO dataset, each image has five captions from five different annotators. Annotators describe the most obvious object or action in an image, which makes this dataset very suitable for the paraphrase generation task. This dataset comes with separate subsets for training and validation: \textit{Train 2014} contains over 82K images and \textit{Val 2014} contains over 40K images. From the five captions accompanying each image, we randomly omit one caption and use the other four as training instances to create paraphrase pairs. In order to compare our results with previous work \cite{prakash2016neural,gupta2017deep}, 20K instances are randomly selected from the data for testing, 10K instances for validation and remaining data over 320K instances for training.

\textbf{Quora} dataset is related to the problem of identifying duplicate questions\footnote{https://data.quora.com/First-Quora-Dataset-Release-Question-Pairs}. It consists of over 400K potential question duplicate pairs. The non-duplicate pairs are related questions or have similar topics, which are not truly semantically equivalent, so we use true examples of duplicate pairs as paraphrase generation dataset. There are a total of 155K such questions. In order to compare our results with previous work \cite{gupta2017deep}, we evaluate our model on 145K training dataset sizes, 5K validation dataset and 4K instances for testing.


\subsection{Evaluation Metric}
To automatically evaluate the performance of paraphrase generation models, we use the well-known evaluation metrics\footnote{We used the evaluation software available at https://github.com/jhclark/multeval} for comparing parallel corpora: BLEU \cite{papineni2002bleu} and METEOR \cite{lavie2007meteor}. Previous work has shown that these metrics can perform well for paraphrase detection \cite{madnani2012re} and correlate well with human judgments in paraphrase generation \cite{wubben2010paraphrase}.

BLEU considers exact matching between reference paraphrases and system generated paraphrases by considering n-gram overlaps. METEOR uses stemming and synonymy in WordNet to improve and smoothen this measure. We report our p-values at $95\%$ Confidence Intervals (CI).

\subsection{Implementation Details}
We leverage the PPDB to build our paraphrased dictionary index and we have introduced our retrieval strategy before. The Paraphrased Database (PPBD)\footnote{http://paraphrase.org} is used to divide the database into six sizes, from \textit{S} up to \textit{XXXL}. We build our paraphrased dictionary index using \textit{L} size PPBD. PPDB contains five types of entailment relations and we exact paraphrased pairs with equivalent entailment relations to ensure the quality of our paraphrased dictionary.

We use NLTK \cite{bird2004nltk} to tokenize the sentences and keep words that appear more than 10 times in our vocabulary. Following the data preprocessing method in previous work \cite{prakash2016neural,gupta2017deep}, we reduce those captions to the size of 15 words (by removing the words beyond the first 15) for the MSCOCO dataset, and sentences whose lengths are greater than 30 words are filtered in the Quora dataset. The max length of phrases in PPDB is set to 7 and the size $M$ of the paraphrased dictionary is 10.

We use a one-hot vector approach to represent the words in all models. The training hyper-parameters are selected based on the results of the validation set. The dimensions of word embeddings is set to 300 and hidden vectors are set to 512 in the sequence encoder and decoder. The dimensions of the attention vector are also set to 512 and the dropout rate is set to 0.5 for regularization. The mini-batched Adam \cite{kingma2014adam} algorithm is used to optimize the objective function. The batch size and base learning rates are set to 64 and 0.001, respectively. 

\subsection{Baselines}
We compare our method with the following baseline methods for paraphrase generation:

\textbf{Seq2Seq}: We implement the standard sequence to sequence with attention model \cite{bahdanau2014neural}, which is implemented in OpenNMT \cite{opennmt}. All the settings are the same as our system.

\textbf{Residual LSTM}: Residual LSTM is a stacked residual LSTM network under the sequence to sequence framework proposed by \cite{prakash2016neural}. It adds residual connections between LSTM layers to help retain essential  words in the generated paraphrases.


\textbf{VAE-SVG}:  VAE-SVG is the current state-of-the-art paraphrase method on the MSCOCO dataset \cite{gupta2017deep}. It combines the variational autoencoder(VAE) and sequence-to-sequence model to generate paraphrases.

\textbf{VAE-SVG-eq}: It is the current state-of-the-art paraphrase method on the Quora dataset \cite{gupta2017deep}. Different from the VAE-SVG model, the encoder of the original sentence is the same on both sides i.e. encoder side and the decoder side in this variation.

\subsection{Evaluation Results}
As shown in Table \ref{results_coco}, we compare our dictionary-guided editing networks model with several state-of-the-art methods on the MSCOCO dataset. The results demonstrate that our model consistently improves performance over other models for both greedy search and beam search. We get further improvement in both metrics though beam search, though, these improvements are not as significant as for Seq2Seq. This could be because the paraphrased dictionary provides some information for generating paraphrases, which could avoid our model to output the paraphrases which have only a few common terms.
For MSCOCO, the comparison between two models is significant at 95\% CI, if the difference in their score is more than $0.2$ in BLEU and $0.1$ in METEOR.

In Table \ref{results_quora}, we report BLEU and METEOR results for the Quora dataset. For
this dataset, we compare the results of our approach with existing approaches at greedy search and beam search. 
The results demonstrate that our proposed model outperforms other models at the non-beam search. For the greedy search, the dictionary-guided editing networks model is able to give a 1.3 performance improvement for BELU and 3.7 improvement for the METEOR metric over the state-of-the-art one. For beam size of 10, our model outperforms other models in the Quora dataset except the VAE-SVG-eq model, in which beam search gives an 11\% absolute point performance improvement in the BLEU score. For the Quora dataset, beam search does not give such a significant improvement in our model. Comparison between two models is significant at 95\% CI, if the difference in their score is more than $0.2$ in BLEU and $0.1$ in METEOR for Quora dataset.

\begin{table}[]
\centering
\caption{Results on MSCOCO dataset. Higher BLEU and METEOR score is better. Scores of the methods marked with * are taken from \cite{gupta2017deep}.}
\label{results_coco}
\begin{tabular}{llll}
\hline
\textbf{Model}  & \textbf{Beam size}   & \textbf{BLEU} & \textbf{METEOR} \\
Seq2Seq         & 1          & 29.9       	  & 24.7		        \\
VAE-SVG*         & 1	         & 39.2           & 29.2                \\
VAE-SVG-eq*      & 1          & 37.3           & 28.5                \\
Our method      & 1          & \textbf{40.3} & \textbf{30.1}  \\ \hline
Seq2Seq*         & 10          & 33.4          & 25.2           \\
Residual LSTM*   & 10          & 37.0          & 27.0           \\
VAE-SVG*         & 10          & 41.3          & 30.9           \\
VAE-SVG-eq*      & 10          & 39.6          & 30.2           \\
Our method      & 10          & \textbf{42.6} & \textbf{31.3}  \\ \hline
\end{tabular}
\end{table}

\begin{table}[]
\centering
\caption{Results on Quora dataset. Higher BLEU and METEOR score is better. Scores of the methods marked with * are taken from \cite{gupta2017deep}.}
\label{results_quora}
\begin{tabular}{llll}
\hline
\textbf{Model}  & \textbf{Beam size} & \textbf{BLEU}  & \textbf{METEOR}     \\ \hline
Seq2Seq         & 1         & 25.9           & 25.8                \\
Residual LSTM   & 1         & 26.3           & 26.2                \\
VAE-SVG*         & 1         & 25.0           & 25.1                \\
VAE-SVG-eq*      & 1         & 26.2           & 25.7                \\
Our method      & 1         & \textbf{27.6}  & \textbf{29.9}      \\ \hline

Seq2Seq         & 10         & 27.9    & 29.3           \\
Residual LSTM   & 10         & 27.4    & 28.9           \\
VAE-SVG-eq*      & 10         & 37.1    & 32.0          \\
Our method      & 10         & 28.4    & 30.6           \\ \hline
\end{tabular}
\end{table}

 \begin{figure}[h]		
			\hspace{1.5mm}	\includegraphics[width=7.19cm,height=4cm]{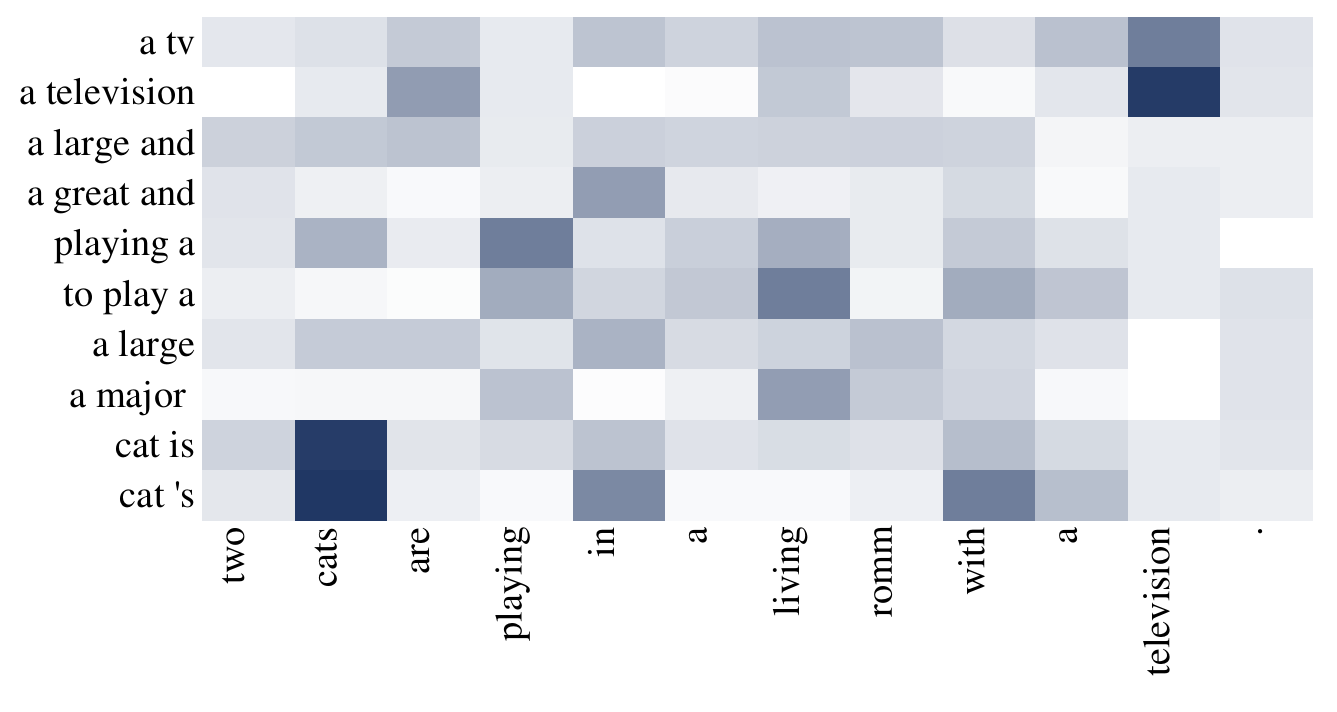}
            \\
            \hspace{2mm}	\includegraphics[width=6.7cm,height=3.78cm]{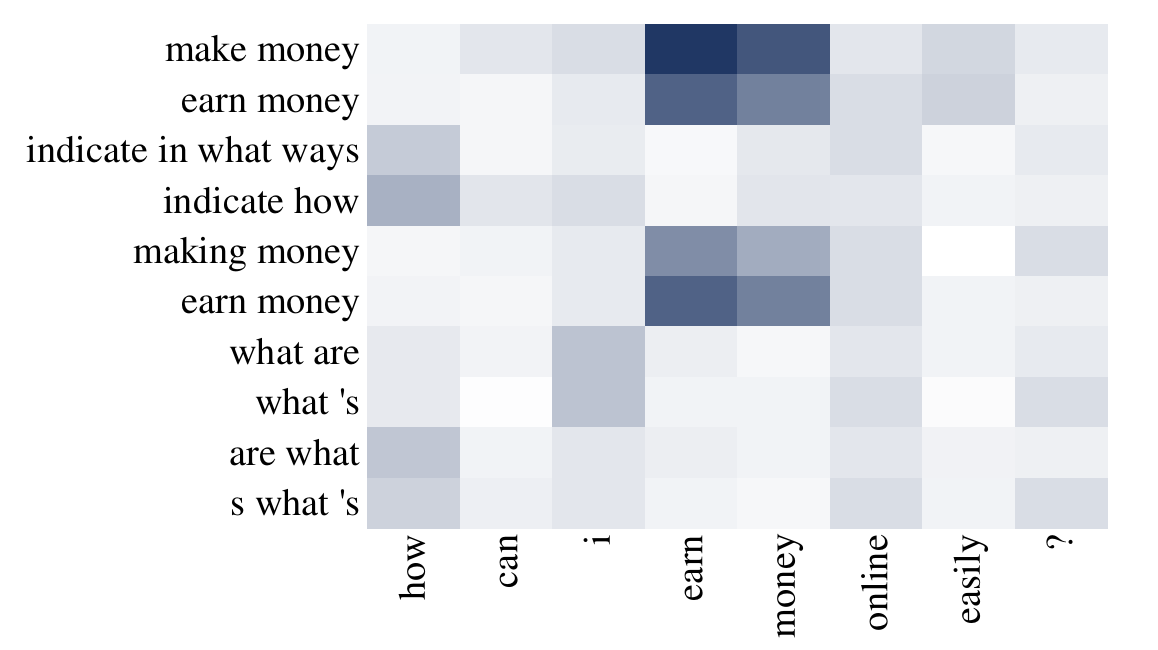}
			\vspace{-2mm}
			\caption{Visualization of dictionary-guided attention in the decoder. Each column in the diagram corresponds to the weights of the decoder and items in the paraphrased dictionary.
			}\label{fig:atten}
		\end{figure}
\subsection{Discussion}

\begin{table}[h]
\centering
\caption{
Example paraphrases generated using the dictionary-guided editing networks on MSCOCO and Quora datasets.}
\label{case_study}

\begin{tabular}{|L{1.5cm}|L{6cm}|}
\hline
Source         	&   these two cats are playing in a room that has \textcolor{blue}{a large tv} and a laptop computer .	\\ \hline 
Reference   	&  a cat being lazy and a cat being nozy in a living room with tv and a laptop displaying the same things .	\\ \hline
Generated       & two cats are playing in a living room with \textcolor{blue}{a television} .	\\ \hline
Dictionary    & \makecell{\textcolor{red}{(a tv, a television)}
						\\ (a large and, a great and)
                        \\ (playing a, to play a)}   \\ \hline
\end{tabular}

\vspace{1mm}

\begin{tabular}{|L{1.5cm}|L{6cm}|}
\hline
Source         	&   a large passenger \textcolor{blue}{airplane} flying through the air .     	\\ \hline 
Reference   	&  an airplane that is , either , landing or just taking off . 	\\ \hline
Generated       & a large \textcolor{blue}{jetliner} flying through a blue sky .					\\ \hline
Dictionary     	& \makecell{(the airplane, the aeroplane)         
						\\ \textcolor{red}{(airplane, jetliner)}
                        \\ (a large, a great)}   \\ \hline
\end{tabular}

\vspace{1mm}

\begin{tabular}{|L{1.5cm}|L{6cm}|}
\hline
Source         &  what are ways i can \textcolor{blue}{make money} online ?           \\ \hline 
Reference   &  can i earn money online ?     \\ \hline
Generated        & how can i \textcolor{blue}{earn money} online easily ?          \\ \hline
Dictionary     &  \makecell{ \textcolor{red}{(make money, earn money)}
						\\ (indicate in what ways, indicate how)
                        \\ \textcolor{red}{(making money, earn money)} }  \\ \hline
\end{tabular}

\vspace{1mm}

\begin{tabular}{|L{1.5cm}|L{6cm}|}
\hline
Source         &  can you \textcolor{blue}{offer} me any advice on how to lose weight ?          \\ \hline 
Reference   & how can i efficiently lose weight ?                    \\ \hline
Generated        & can you \textcolor{blue}{give} me some advice on losing weight ?                \\ \hline
Dictionary     & \makecell{(offer advice, provide advice)         
						\\ \textcolor{red}{(offer advice, give advice)}
                        \\ (you lost weight, you 've lost weight)}           \\ \hline
\end{tabular}
\end{table}

In Figure~\ref{fig:atten}, we show the visualization of dictionary-guided attention in the decoder. Each column in the diagram corresponds to the weights of the decoder and items in the paraphrased dictionary.

Figure \ref{fig:atten} shows two examples separately from MSCOCO and Quora datasets. Each example has five paraphrased pairs in the dictionary. The delete attention and insert attention scores are represented by gray scales and are column-wisely normalized as described in Equation \ref{eq:del} and \ref{eq:ins}. As described, the editing attention mechanism learns soft alignment scores between paraphrased dictionary and generated words. These scores are used to guide the decision for word deletion or insertion in the decoder. 

In the first example, the generated paraphrase is \textit{"two cats are playing in a living room with a television ."}. We find that the pair (\textit{"a tv"}, \textit{"a television"}) has larger attention scores where the decoder generates the \textit{television} word. This demonstrates our paraphrased dictionary has more effect on generating some words which might be deleted or inserted. As we can see in the second example, the model learns alignments when the decoder generates \textit{earn money}.

In Table~\ref{case_study}, we show some generated paraphrase examples on MSCOCO and Quora datasets.  In these tables, red denotes paraphrased dictionary pairs which might be used to guide paraphrase generation and blue denotes phrases which are found in the paraphrased dictionary. As we can see, our model is able to replace some words or phrases in the original sentence based on the dictionary and makes necessary changes to ensure the new sentence is grammatically correct and fluent.


\section{Conclusion}
In this paper, we present a dictionary-guided editing networks model for generating paraphrase sentences through editing the original sentence. It can effectively leverage word level and phrase level paraphrase pairs from an off-the-shelf dictionary. The system jointly learns the selection of the appropriate word level and phrase level paraphrase pairs in the context of the original sentence from the Paraphrase Database (PPDB) as well as the generation of fluent natural language sentences. Experiments on the Quora and MSCOCO datasets demonstrate that the dictionary-guided editing networks significantly improves the existing generative models for paraphrase generation from scratch. The dictionary-guided editing networks can also be applied  to other text generation tasks, such as the text style transfer where we can use word and phrase level style mapping dictionaries to facilitate sentence level style transfer results.

			\small
		\bibliographystyle{aaai}
		\bibliography{acl2017}

\begin{thebibliography}{}

\bibitem[\protect\citeauthoryear{Bahdanau, Cho, and
  Bengio}{2015}]{bahdanau2014neural}
Bahdanau, D.; Cho, K.; and Bengio, Y.
\newblock 2015.
\newblock Neural machine translation by jointly learning to align and
  translate.
\newblock {\em ICLR}.

\bibitem[\protect\citeauthoryear{Bird and Loper}{2004}]{bird2004nltk}
Bird, S., and Loper, E.
\newblock 2004.
\newblock Nltk: the natural language toolkit.
\newblock In {\em Proceedings of the ACL 2004 on Interactive poster and
  demonstration sessions}, ~31.
\newblock Association for Computational Linguistics.

\bibitem[\protect\citeauthoryear{Cao \bgroup et al\mbox.\egroup
  }{2018}]{ziqiang2018summary}
Cao, Z.; Li, W.; Wei, F.; and Li, S.
\newblock 2018.
\newblock Retrieve, rerank and rewrite: Soft template based neural
  summarization.
\newblock Association for Computational Linguistics.

\bibitem[\protect\citeauthoryear{Ganitkevitch, Van~Durme, and
  Callison-Burch}{2013}]{ganitkevitch2013ppdb}
Ganitkevitch, J.; Van~Durme, B.; and Callison-Burch, C.
\newblock 2013.
\newblock Ppdb: The paraphrase database.
\newblock In {\em Proceedings of the 2013 Conference of the North American
  Chapter of the Association for Computational Linguistics: Human Language
  Technologies},  758--764.

\bibitem[\protect\citeauthoryear{Gupta \bgroup et al\mbox.\egroup
  }{2017}]{gupta2017deep}
Gupta, A.; Agarwal, A.; Singh, P.; and Rai, P.
\newblock 2017.
\newblock A deep generative framework for paraphrase generation.
\newblock {\em arXiv preprint arXiv:1709.05074}.

\bibitem[\protect\citeauthoryear{Guu \bgroup et al\mbox.\egroup
  }{2017}]{DBLP:journals/corr/abs-1709-08878}
Guu, K.; Hashimoto, T.~B.; Oren, Y.; and Liang, P.
\newblock 2017.
\newblock Generating sentences by editing prototypes.
\newblock {\em TACL}.

\bibitem[\protect\citeauthoryear{Hassan \bgroup et al\mbox.\egroup
  }{2007}]{hassan2007unt}
Hassan, S.; Csomai, A.; Banea, C.; Sinha, R.; and Mihalcea, R.
\newblock 2007.
\newblock Unt: Subfinder: Combining knowledge sources for automatic lexical
  substitution.
\newblock In {\em Proceedings of the 4th International Workshop on Semantic
  Evaluations},  410--413.
\newblock Association for Computational Linguistics.

\bibitem[\protect\citeauthoryear{Kingma and Ba}{2014}]{kingma2014adam}
Kingma, D.~P., and Ba, J.
\newblock 2014.
\newblock Adam: A method for stochastic optimization.
\newblock {\em arXiv preprint arXiv:1412.6980}.

\bibitem[\protect\citeauthoryear{Klein \bgroup et al\mbox.\egroup
  }{2017}]{opennmt}
Klein, G.; Kim, Y.; Deng, Y.; Senellart, J.; and Rush, A.~M.
\newblock 2017.
\newblock Opennmt: Open-source toolkit for neural machine translation.
\newblock In {\em Proc. ACL}.

\bibitem[\protect\citeauthoryear{Kozlowski, McCoy, and
  Vijay-Shanker}{2003}]{kozlowski2003generation}
Kozlowski, R.; McCoy, K.~F.; and Vijay-Shanker, K.
\newblock 2003.
\newblock Generation of single-sentence paraphrases from predicate/argument
  structure using lexico-grammatical resources.
\newblock In {\em Proceedings of the second international workshop on
  Paraphrasing-Volume 16},  1--8.
\newblock Association for Computational Linguistics.

\bibitem[\protect\citeauthoryear{Lavie and Agarwal}{2007}]{lavie2007meteor}
Lavie, A., and Agarwal, A.
\newblock 2007.
\newblock Meteor: An automatic metric for mt evaluation with high levels of
  correlation with human judgments.
\newblock In {\em Proceedings of the Second Workshop on Statistical Machine
  Translation},  228--231.
\newblock Association for Computational Linguistics.

\bibitem[\protect\citeauthoryear{Li \bgroup et al\mbox.\egroup
  }{2018}]{li2018delete}
Li, J.; Jia, R.; He, H.; and Liang, P.
\newblock 2018.
\newblock Delete, retrieve, generate: A simple approach to sentiment and style
  transfer.
\newblock {\em NAACL}.

\bibitem[\protect\citeauthoryear{Lin \bgroup et al\mbox.\egroup
  }{2014}]{lin2014microsoft}
Lin, T.-Y.; Maire, M.; Belongie, S.; Hays, J.; Perona, P.; Ramanan, D.;
  Doll{\'a}r, P.; and Zitnick, C.~L.
\newblock 2014.
\newblock Microsoft coco: Common objects in context.
\newblock In {\em European conference on computer vision},  740--755.
\newblock Springer.

\bibitem[\protect\citeauthoryear{Luong, Pham, and
  Manning}{2015}]{luong2015effective}
Luong, M.-T.; Pham, H.; and Manning, C.~D.
\newblock 2015.
\newblock Effective approaches to attention-based neural machine translation.
\newblock {\em arXiv preprint arXiv:1508.04025}.

\bibitem[\protect\citeauthoryear{Madnani and
  Dorr}{2010}]{madnani2010generating}
Madnani, N., and Dorr, B.~J.
\newblock 2010.
\newblock Generating phrasal and sentential paraphrases: A survey of
  data-driven methods.
\newblock {\em Computational Linguistics} 36(3):341--387.

\bibitem[\protect\citeauthoryear{Madnani, Tetreault, and
  Chodorow}{2012}]{madnani2012re}
Madnani, N.; Tetreault, J.; and Chodorow, M.
\newblock 2012.
\newblock Re-examining machine translation metrics for paraphrase
  identification.
\newblock In {\em Proceedings of the 2012 Conference of the North American
  Chapter of the Association for Computational Linguistics: Human Language
  Technologies},  182--190.
\newblock Association for Computational Linguistics.

\bibitem[\protect\citeauthoryear{Papineni \bgroup et al\mbox.\egroup
  }{2002}]{papineni2002bleu}
Papineni, K.; Roukos, S.; Ward, T.; and Zhu, W.-J.
\newblock 2002.
\newblock Bleu: a method for automatic evaluation of machine translation.
\newblock In {\em Proceedings of the 40th annual meeting on association for
  computational linguistics},  311--318.
\newblock Association for Computational Linguistics.

\bibitem[\protect\citeauthoryear{Pavlick \bgroup et al\mbox.\egroup
  }{2015}]{pavlick2015ppdb}
Pavlick, E.; Rastogi, P.; Ganitkevitch, J.; Van~Durme, B.; and Callison-Burch,
  C.
\newblock 2015.
\newblock Ppdb 2.0: Better paraphrase ranking, fine-grained entailment
  relations, word embeddings, and style classification.
\newblock In {\em Proceedings of the 53rd Annual Meeting of the Association for
  Computational Linguistics and the 7th International Joint Conference on
  Natural Language Processing (Volume 2: Short Papers)}, volume~2,  425--430.

\bibitem[\protect\citeauthoryear{Prakash \bgroup et al\mbox.\egroup
  }{2016}]{prakash2016neural}
Prakash, A.; Hasan, S.~A.; Lee, K.; Datla, V.; Qadir, A.; Liu, J.; and Farri,
  O.
\newblock 2016.
\newblock Neural paraphrase generation with stacked residual lstm networks.
\newblock {\em arXiv preprint arXiv:1610.03098}.

\bibitem[\protect\citeauthoryear{Quirk, Brockett, and
  Dolan}{2004}]{quirk2004monolingual}
Quirk, C.; Brockett, C.; and Dolan, B.
\newblock 2004.
\newblock Monolingual machine translation for paraphrase generation.

\bibitem[\protect\citeauthoryear{Wubben, Van Den~Bosch, and
  Krahmer}{2010}]{wubben2010paraphrase}
Wubben, S.; Van Den~Bosch, A.; and Krahmer, E.
\newblock 2010.
\newblock Paraphrase generation as monolingual translation: Data and
  evaluation.
\newblock In {\em Proceedings of the 6th International Natural Language
  Generation Conference},  203--207.
\newblock Association for Computational Linguistics.

\bibitem[\protect\citeauthoryear{Zaremba, Sutskever, and
  Vinyals}{2014}]{zaremba2014recurrent}
Zaremba, W.; Sutskever, I.; and Vinyals, O.
\newblock 2014.
\newblock Recurrent neural network regularization.
\newblock {\em arXiv preprint arXiv:1409.2329}.

\bibitem[\protect\citeauthoryear{Zhao \bgroup et al\mbox.\egroup
  }{2008}]{zhao2008combining}
Zhao, S.; Niu, C.; Zhou, M.; Liu, T.; and Li, S.
\newblock 2008.
\newblock Combining multiple resources to improve smt-based paraphrasing model.
\newblock {\em Proceedings of ACL-08: HLT}  1021--1029.

\bibitem[\protect\citeauthoryear{Zhao \bgroup et al\mbox.\egroup
  }{2009}]{zhao2009application}
Zhao, S.; Lan, X.; Liu, T.; and Li, S.
\newblock 2009.
\newblock Application-driven statistical paraphrase generation.
\newblock In {\em Proceedings of the Joint Conference of the 47th Annual
  Meeting of the ACL and the 4th International Joint Conference on Natural
  Language Processing of the AFNLP: Volume 2-Volume 2},  834--842.
\newblock Association for Computational Linguistics.

\bibitem[\protect\citeauthoryear{Zhao \bgroup et al\mbox.\egroup
  }{2010}]{zhao2010leveraging}
Zhao, S.; Wang, H.; Lan, X.; and Liu, T.
\newblock 2010.
\newblock Leveraging multiple mt engines for paraphrase generation.
\newblock In {\em Proceedings of the 23rd International Conference on
  Computational Linguistics},  1326--1334.
\newblock Association for Computational Linguistics.

\end{thebibliography}
	\end{CJK*}
\end{document}